\documentclass[11pt,a4paper]{article}

\usepackage[hyperref]{acl2021}
\usepackage{times}
\usepackage{latexsym}

\usepackage{url}

\usepackage{caption}
\usepackage{amsmath}
\usepackage[ruled,linesnumbered]{algorithm2e} 
\usepackage{amssymb}
\usepackage{bm}
\usepackage{empheq}
\usepackage{adjustbox}
\usepackage{tabulary} 
\usepackage{multirow}
\usepackage{makecell}
\usepackage{color} 
\usepackage{setspace} 

\usepackage{xcolor}
\definecolor{green_}{HTML}{DDF4EA}
\definecolor{blue_}{HTML}{DEEFF9}
\definecolor{pink_}{HTML}{FFF0DD}
\definecolor{grey_}{HTML}{C4CBCF}
\usepackage[normalem]{ulem} 
\newcommand\hlgreen{\bgroup\markoverwith
  {\textcolor{green_}{\rule[-.5ex]{2pt}{2.5ex}}}\ULon}
\newcommand\hlblue{\bgroup\markoverwith
  {\textcolor{blue_}{\rule[-.5ex]{2pt}{2.5ex}}}\ULon}
\newcommand\hlpink{\bgroup\markoverwith
  {\textcolor{pink_}{\rule[-.5ex]{2pt}{2.5ex}}}\ULon}
\newcommand\hlgrey{\bgroup\markoverwith
  {\textcolor{grey_}{\rule[-.5ex]{2pt}{2.5ex}}}\ULon}

\usepackage{pbox}
\usepackage{graphicx, subfigure}
\usepackage{booktabs}
\usepackage{tikz}
\usetikzlibrary{fit,positioning}
\usetikzlibrary{arrows}
\usepackage{etoolbox}
\newcommand{\circled}[2][]{\tikz[baseline=(char.base)]
    {\node[shape = circle, draw, inner sep = 0.0pt]
    (char) {\phantom{\ifblank{#1}{#2}{#1}}};%
    \node at (char.center) {\makebox[0pt][c]{#2}};}}
\robustify{\circled}
\newcommand{\tabincell}[2]{\begin{tabular}{@{}#1@{}}#2\end{tabular}}  

\aclfinalcopy 
\def\aclpaperid{1915} 

\title{
  From Paraphrasing to Semantic Parsing: Unsupervised Semantic Parsing via Synchronous Semantic Decoding
}

\author{
  Shan Wu$^{1,3}$,
  Bo Chen$^{1}$,
  Chunlei Xin$^{1,3}$,  
  Xianpei Han$^{1,2,}$\footnotemark[1] $^{\,}$,
  Le Sun$^{1,2,}$\thanks{~ Corresponding Author} $^{\,}$,\\
  \textbf{Weipeng Zhang$^{4}$, 
      Jiansong Chen$^{4}$,
      Fan Yang$^{4}$,
      Xunliang Cai$^{4}$}  \\
  $^{1}$Chinese Information Processing Laboratory ~ $^{2}$State Key Laboratory of Computer Science \\
   Institute of Software, Chinese Academy of Sciences, Beijing, China \\
  $^{3}$University of Chinese Academy of Sciences, Beijing, China ~ $^{4}$Meituan\\
  \tt{\{wushan2018,chenbo,xianpei,sunle\}@iscas.ac.cn}, \\
  \tt{xinchunlei20@mails.ucas.ac.cn},\\
  \tt{\{zhangweipeng02,chenjiansong,yangfan79,caixunliang\}@meituan.com}}

\begin{document}
\date{2021-5-26 14:08:30}

\maketitle

\begin{abstract}

Semantic parsing is challenging due to the structure gap and the semantic gap between utterances and logical forms. 
In this paper, we propose an unsupervised semantic parsing method -- Synchronous Semantic Decoding (SSD), 
which can simultaneously resolve the semantic gap and the structure gap by jointly leveraging paraphrasing and grammar-constrained decoding. 
Specifically, we reformulate semantic parsing as a constrained paraphrasing problem: 
given an utterance, our model synchronously generates its canonical utterance$\footnote{Canonical utterances are  pseudo-language representations of logical forms, which have the synchronous structure of logical forms.\citep{Berant:2014,Chunyang:2016,Su:2017,Cao:2020}}$ and meaning representation. 
During synchronous decoding: the utterance paraphrasing is constrained by the structure of the logical form, therefore the canonical utterance can be paraphrased controlledly; 
the semantic decoding is guided by the semantics of the canonical utterance, therefore its logical form can be generated unsupervisedly. 
Experimental results show that SSD is a promising approach and can achieve competitive unsupervised semantic parsing performance on multiple datasets.








\end{abstract}

\section{Introduction}

Semantic parsing aims to translate natural language utterances to their formal meaning representations, such as lambda calculus \citep{Luke:2005,Wong:2007}, FunQL \citep{Kate:2005,WeiLu:2008}, and SQL queries. 
Currently, most neural semantic parsers \citep{Dong:2016,Chen:2018,Zhao:2020,Shao:2020} model semantic parsing as a sequence to sequence translation task via encoder-decoder framework. 

\begin{figure}[t]
  \centering
  \noindent\makebox[.45\textwidth][r] {
  \includegraphics[width=.45\textwidth]{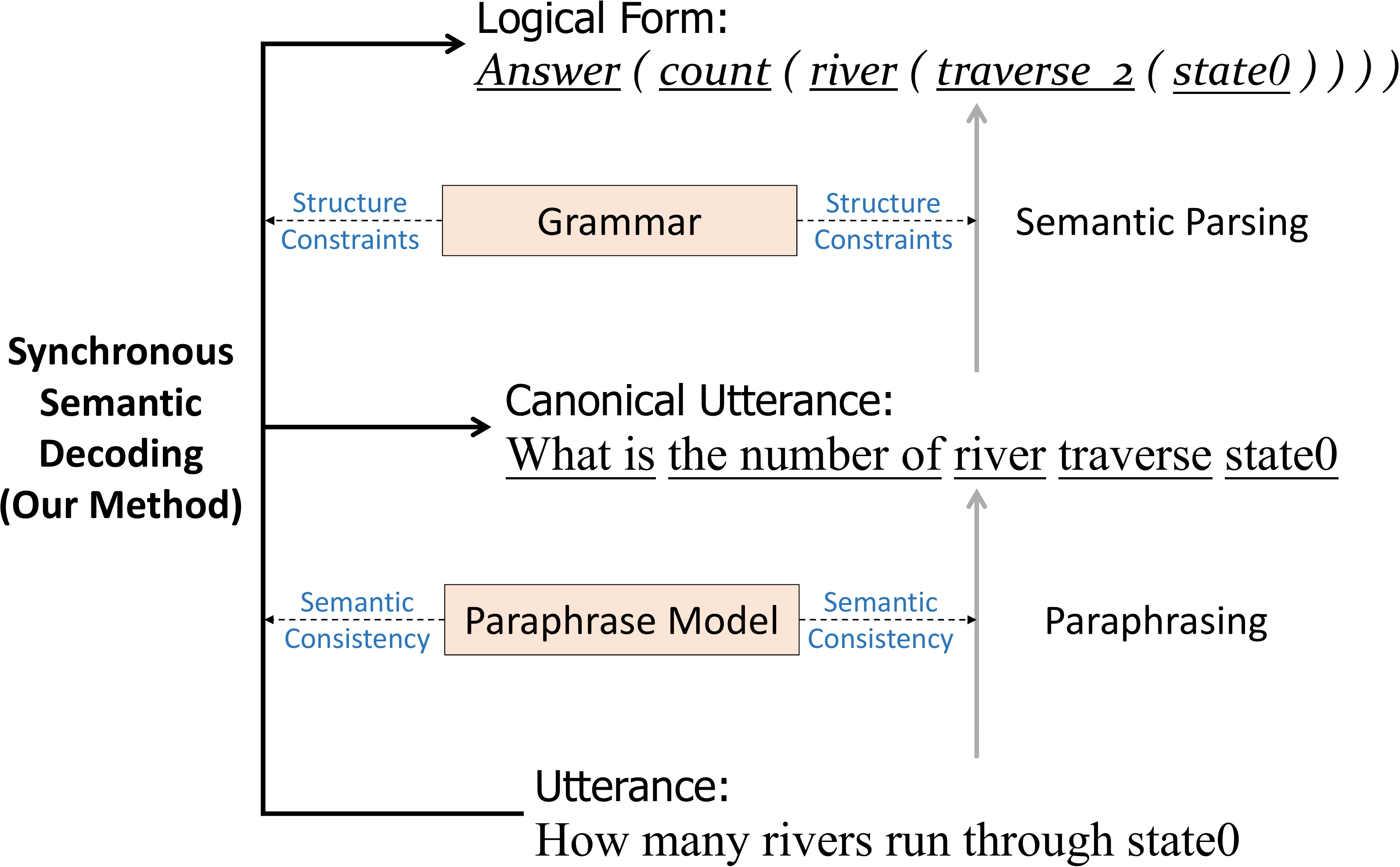}
  }
      \setlength{\abovecaptionskip}{0.3cm}
    \setlength{\belowcaptionskip}{-0.5cm}
  \caption{
  \small{
    Different from previous staged methods (indicated by gray lines), our method generates canonical utterance and logical form synchronously.
    The semantic gap and the structure gap are simultaneously resolved by jointly leveraging paraphrasing and grammar-constrained decoding. Thus, our synchronous decoding employs both the semantic and the structure constraints to solve unsupervised semantic parsing.
    }
  }
  \label{fig1} 
\end{figure}

Semantic parsing is a challenging task due to the structure gap and the semantic gap between natural language utterances and logical forms. 
For structure gap, because utterances are usually word sequences and logical forms are usually trees/graphs constrained by specific grammars, a semantic parser needs to learn the complex structure transformation rules between them. 
For semantic gap, because the flexibility of natural languages, the same meaning can be expressed using very different utterances, a semantic parser needs be able to map various expressions to their semantic form. 
To address the structure gap and the semantic gap, current semantic parsers usually rely on a large amount of labeled data, often resulting in data bottleneck problem.

Previous studies have found that the structure gap and the semantic gap can be alleviated by leveraging external resources, 
therefore the reliance on data can be reduced. 
For structure gap, 
previous studies found that constrained decoding can effectively constrain the output structure 
by injecting grammars of logical forms and facts in knowledge bases during inference. 
For example, the grammar-based neural semantic parsers \citep{Chunyang:2016,Yin:2017}
and the constrained decoding algorithm \citep{Krishnamurthy:2017}. 
For semantic gap, 
previous studies have found that paraphrasing is an effective technique for resolving the diversity of natural expressions.
Using paraphrasing, 
semantic parsers can handle the different expressions of the same meaning,
therefore can reduce the requirement of labeled data. 
For example, supervised methods \citep{Berant:2014,Su:2017} use the paraphrasing scores between canonical utterances and sentences to re-rank logical forms; 
Two-stage \citep{Cao:2020} rewrites utterances to canonical utterances which can be easily parsed. 
The main drawback of these studies is that they use constrained decoding and paraphrasing independently and separately, therefore they can only alleviate either semantic gap or structure gap.

In this paper, we propose an unsupervised semantic parsing method -- \textit{Synchronous Semantic Decoding (SSD)}, which can simultaneously resolve the structure gap and the semantic gap by jointly leveraging paraphrasing and grammar-constrained decoding. Specifically, we model semantic parsing as a constrained paraphrasing task: given an utterance, we synchronously decode its canonical utterance and its logical form using a general paraphrase model, where the canonical utterance and the logical form share the same underlying structure. 
Based on the synchronous decoding, the canonical utterance generation can be constrained by the structure of logical form, and the logical form generation can be guided
by the semantics of canonical form.
By modeling the interdependency between canonical utterance and logical form, and exploiting them through synchronous decoding, our method can perform effective unsupervised semantic parsing using only pretrained general paraphrasing model – no annotated data for semantic parsing is needed.

We conduct experiments on \textsc{Geo} and \textsc{Overnight}. 
Experimental results show that our method is promising, which can achieve competitive unsupervised semantic parsing performance, and can be further improved with external resources. 
The main contributions of this paper are:
\begin{itemize}
  \item  We propose an unsupervised semantic parsing method -- Synchronous Semantic Decoding , which can simultaneously resolve the semantic gap and the structure gap by jointly leveraging paraphrasing and grammar-constrained semantic decoding.
  \item  We design two effective synchronous semantic decoding algorithms -- rule-level inference and word-level inference, which can generate paraphrases under the grammar constraints and synchronously decode meaning representations.
  \item  Our model achieves competitive unsupervised semantic parsing performance on \textsc{Geo} and \textsc{Overnight} datasets.
\end{itemize}


\begin{figure*}[!t]
  \centering
  \includegraphics[width=.8\textwidth]{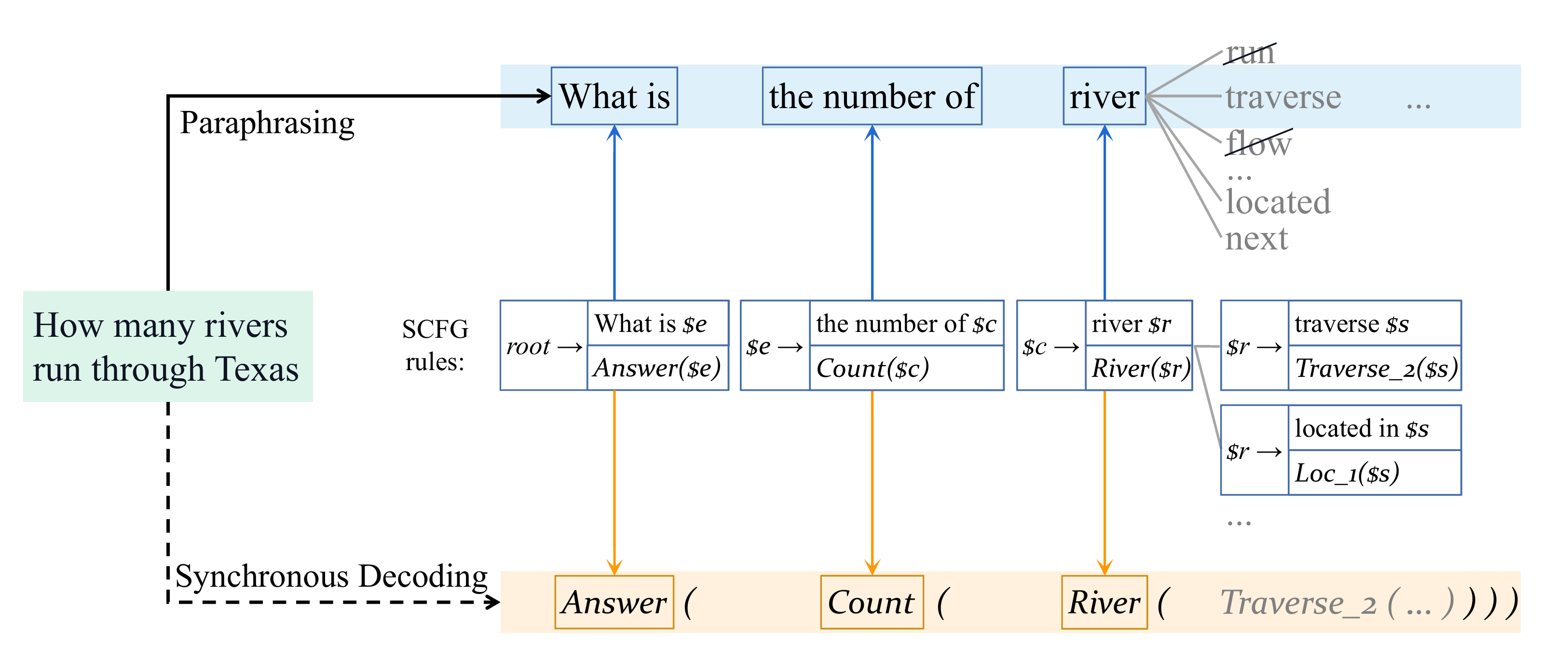} 
      \setlength{\abovecaptionskip}{0.3cm}
    \setlength{\belowcaptionskip}{-0.3cm}
  \caption{ \small{Overview of our approach. \hlgreen{The sentence} is paraphrased to \hlblue{canonical utterance} and parsed to \hlpink{logical form} synchronously. When decoding ``traverse'', the paraphrase model tends to generate the words such as ``run'', ``flow'', ``traverse'' to preserve semantics. And synchronous grammar limits the next words of the canonical utterance to follow the candidate production rules. Then it is easy to discard ``run'' and ``flow'', and select the most likely word ``traverse'' with its production rule from the candidates. In this parper, we propose rule-level and word-level inference methods to decode words and production rules synchronously.}} 
  \label{figbig} 
\end{figure*}

\section{Model Overview}

We now present overview of our synchronous semantic decoding algorithm, which can jointly leverage paraphrasing and grammar-constrained decoding for unsupervised semantic parsing. Given an utterance, SSD reformulates semantic parsing as a constrained paraphrasing problem, and synchronously generates its canonical utterance and logical form. For example in Fig. \ref{figbig}, given ``How many rivers run through Texas'', SSD generates ``What is the number of river traverse State0'' as its canonical form and \texttt{Answer(Count(River(Traverse\_2(\\
State0))))} as its logical form. During synchronous decoding: 
the utterance paraphrase generation is constrained by the grammar of logical forms, 
therefore the canonical utterance can be generated controlledly; 
the logical form is generated synchronously with the canonical utterance via synchronous grammar. 
Logical form generation is controlled by the semantic constraints from paraphrasing and structure constraints from grammars and database schemas. 
Therefore the logical form can be generated unsupervisedly.

To this end, SSD needs to address two challenges. 
Firstly, we need to design paraphrasing-based decoding algorithms which can effectively impose grammar constraints on inference. 
Secondly, current paraphrasing models are trained on natural language sentences, which are different from the unnatural canonical utterances.
Therefore SSD needs to resolve this style bias for effective canonical utterance generation.

Specifically, we first propose two inference algorithms for constrained paraphrasing based synchronous semantic decoding: 
rule-level inference and word-level inference. 
Then we resolve the style bias of paraphrase model 
via adaptive fine-tuning and utterance reranking, 
where adaptive fine-tuning can adjust the paraphrase model to generate canonical utterances, 
and utterance reranking resolves the style bias by focusing more on semantic coherence. 
In Sections 3-5,
we provide the details of our implementation. 



\section{Synchronous Semantic Decoding}

Given an utterance $x$, we turn semantic parsing into a constrained paraphrasing task. Concretely, we use synchronous context-free grammar as our synchronous grammar, which provides a one-to-one mapping from a logical form $y$ to its canonical utterance $c^y$. The parsing task $\hat{y} = \arg\max\limits_{y \in \mathcal{Y}} p_{\text{\footnotesize{parse}}}(y|x)$ is then transferred to $\hat{y} = \arg\max\limits_{y \in \mathcal{Y}} p_{\text{\footnotesize{paraphrase}}}(c^y|x)$.  
Instead of directly parsing utterance into its logical form, SSD generates its canonical utterance and obtains its logical form based on the one-to-one mapping relation.
In following we first introduce the grammar constraints in decoding, and then present two inference algorithms for generating paraphrases under the grammar constraints.

\subsection{Grammar Constraints in Decoding} \label{constraints}

Synchronous context-free grammar(SCFG) is employed as our synchronous grammar, which is widely used to convert a meaning representation into an unique canonical utterance \citep{Wang:2015,Robin:2016}. 
An SCFG consists of a set of production rules:
$N \rightarrow \left< \alpha,\beta \right>$, where $N$ is a non-terminal, and $\alpha$ and $\beta$ are
sequence of terminal and non-terminal symbols. Each non-terminal symbol in $\alpha$ is aligned to the same non-terminal
symbol in $\beta$, and vice versa. Therefore, an SCFG defines a
set of joint derivations of aligned pairs of utterances and logical forms. 

SCFGs can provide useful constraints for semantic decoding by restricting the decoding space and exploiting the semantic knowledge: 

\paragraph{Grammar Constraints} The grammars ensure the generated utterances/logical forms are grammar-legal. In this way the search space can be greatly reduced. For example,  when expanding the non-terminal \texttt{\$r} in Fig \ref{figbig}
we don't need to consider the words ``run'' and ``flow'', because they are not in the candidate grammar rules.

\paragraph{Semantic Constraints} Like the  type checking in \citet{Wang:2015}, the constraints of knowledge base schema can be integrated to further refine the grammar. 
The semantic constraints ensure the generated utterances/logical forms will be semantically valid. 

\subsection{Decoding}

\subsubsection{Rule-Level Inference}

\begin{figure*}[!t]
  %
  
  \centering
\subfigure[Rule-Level Inference]{
  \begin{minipage}[t]{0.233\textwidth}
    \centering
    \includegraphics[width=0.98\textwidth]{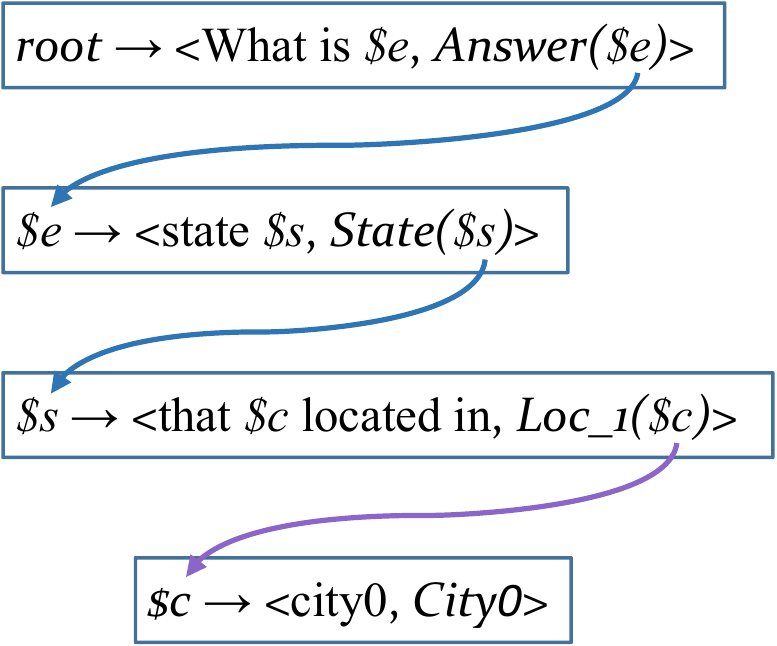} 
  \end{minipage}
}
\subfigure[Word-Level Inference]{
  \begin{minipage}[t]{0.50\textwidth}
  $\ \ \ \ \ $
  \includegraphics[width=0.98\textwidth]{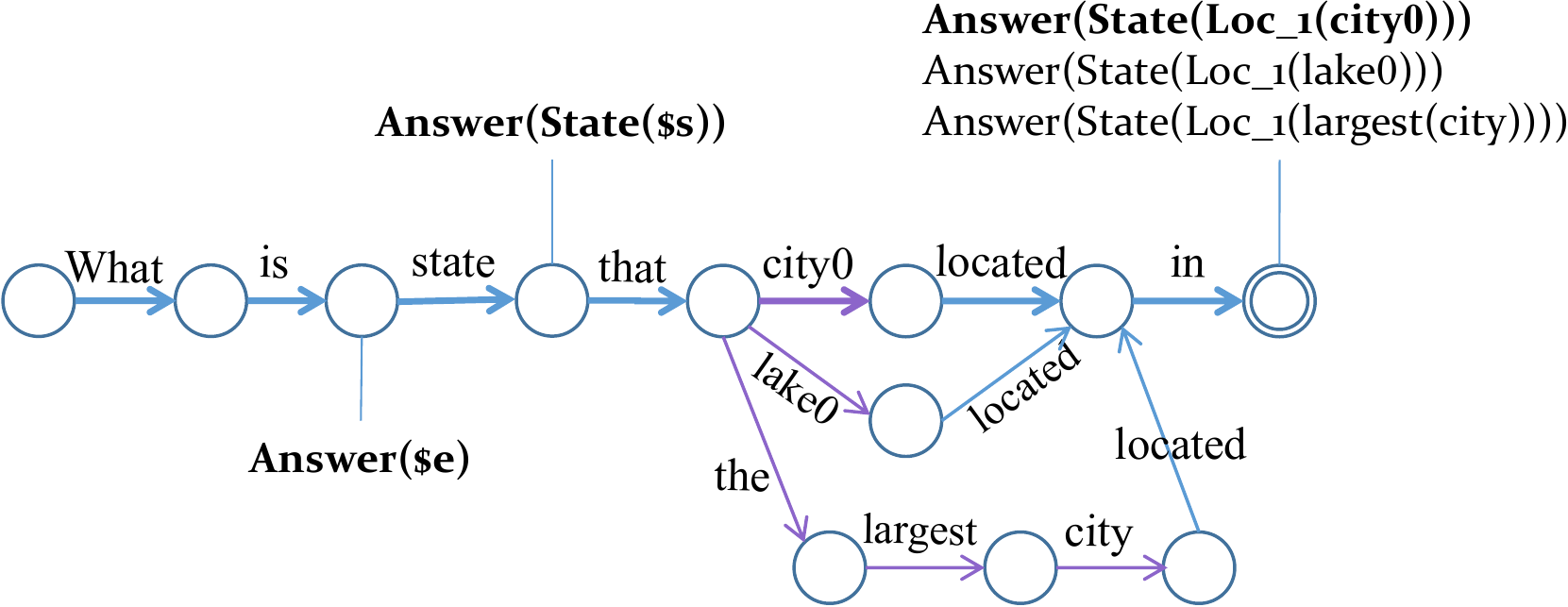} 
  \end{minipage}%
}
  \caption{From the utterance ``\textit{which state is city0 in}'', two inference methods generate its canonical utterance ``\textit{what is state that city0 located in}'' and its logical form \texttt{Answer(State(Loc\_1(City0)))}.  The ways they handle non-terminal \texttt{\$c} which is not at the end of utterance-side production rule are represented by purple lines.}
  \label{two-level}
\end{figure*}



\newcommand{\hangin}{\hangafter 1  \setlength{\hangindent}{\hangindent+1.2em}}
\newcommand{\hanginm}{\hangafter 1  \setlength{\hangindent}{\hangindent-3em}}
\begin{algorithm}[ht]
 \SetKwInOut{KIN}{Input}
 \SetKwInOut{KOUT}{Output}
\small
 \caption{Rule-level inference}
 \label{alg:rule}

 \KIN{\hanginm input utterance $x$, paraphrasing model $Para$, beam size $B$, maximum output length $L$, SCFG rules $R$, maximum search depth $K$;
       }
\SetInd{0.2em}{0.55em}
\SetAlgoNoEnd 


      \hangin {$beam_0 \leftarrow \{\left<s\right>\}$} 

      \hangin {$outputs \leftarrow \{\}$} 

      \For{$t=1$ \KwTo $L$}  
      {
        \For{${\rm hypothesis}$ $c$ \bf{in} $beam_{t-1}$ } {
          \For{$r$ \bf{in} {\rm expand rules for} $c$ } {
            \eIf{{\rm all non-terminals in} $r_\beta$ {\rm are on the right}}{
              $c' \leftarrow \text{Expand}(c,r)$


              $beam_{t} \leftarrow beam_{t} \cup \{c'\} $ 

            }{
              $c' \leftarrow \text{expand}(c,r)$

              $beam_t^{c'} \leftarrow \{c'\}$ 

              \For{$k=1$ \KwTo $K$}{
                \For{${\rm hypothesis}$ $h$ \bf{in} $beam_{t+k-1}^{c'}$}{
                  \hangin{$r_h$ $\leftarrow$ expand rules for $h$'s first non-terminal}

                  $beam_{_{t\!+\!k\!}}^{c'}\leftarrow\!beam_{_{t\!+\!k}}^{c'}\!\!\cup\text{Expand}(h,r_h)$

                }
                $beam_{_{t\!+\!k\!}}^{c'} \leftarrow \text{NBest}(beam_{_{t\!+\!k\!}}^{c'},B)$

                \hangin{Move utterances from $beam_{_{t\!+\!k\!}}^{c'}$ to $beam_{_{t\!+\!k\!}}$, if non-terminals are on the right of the utterances.}   
              }
            }
          }
        }
        \hangin {$beam_{t} \leftarrow$ NBest($beam_{t}$,$B-|outputs|$)}

        \hangin Move full utterances from $beam_{t}$ to $outputs$

        \If{$beam_{t}$ is empty}
        {\Return{$outputs$}}
      }
    \Return{$outputs$}
\end{algorithm}

One strategy to generate paraphrase under the grammar constraint is taking the grammar rule as the decoding unit. Grammar-based decoders have been proposed to output sequences of grammar rules instead of words\citep{Yin:2017}. Like them, our rule-level inference method takes the grammar rule as the decoding unit. Figure \ref{two-level} (a) shows an example of our rule level inference method. 

When the non-terminal in the utterance-side production rule is at the end of the rule 
(e.g., $\text{\$e}\rightarrow\left<\text{state \$s}, \texttt{State(}\text{\$s}\texttt{)}\right>$),
denoting the utterance-side production rule as $r_{\beta}=[w_{1},w_{2},...,w_{L_r},N]$, 
we can simply expand non-terminals in canonical utterances by this rule,  
and generate the canonical utterances from left to right with probabilities computed by:

\begin{small}
\begin{equation}
  \begin{aligned}
P(c^{y_{\le t}}|x) = P(c^{y_{< t}}|x) \prod_{i=1}^{L_r} P_{\small{\text{paraphrase}}}(w_i|x,c^{y_{< t}},w_{<i})
\label{commonrule}
  \end{aligned}
\end{equation}
\end{small}
Otherwise, we generate the next production rules to expand this rule (i.e., rule with purple line), until 
there is no non-terminal on the left of words, or the generating step reaches the depth of $K$. We use beam search during the inference. 
The inference details are described in Algorithm \ref{alg:rule}.


\subsubsection{Word-Level Inference}




Except for rule-level inference, we also propose a word-level inference algorithm, which 
generates paraphrases word by word under the SCFG constraints.


Firstly, we construct a deterministic automaton using LR(1) parser \citep{Knuth:1965} from the CFG in utterance side.
The automaton can transit from one state to another in response to an input. 
The inputs of the automaton are words and the states of it are utterance/logical form segments. 
LR(1) parser peeks ahead one lookahead input symbol, and the state transition table describes the acceptable inputs
and the next states. 

Then, in each decoding step we generate a word with a new state which is transited from previous state. 
An example is shown in Figure \ref{two-level} (b).
Only the acceptable words in the current state can be generated, and the end-of-sentence symbol can only be generated when reaching the final state. 
Beam search is also used in this inference.







\section{Adaptive Fine-tuning} \label{adaptive}


The above decoding algorithms only rely on a paraphrase generation model 
, which generates canonical utterance and logical form synchronously
for semantic parsing. We can directly use general paraphrase generation models such as GPT-2\citep{Radford:2018}, T5\citep{Raffel:2020} for SSD. 
However, as described in above, there exists a style bias between natural language sentences and canonical utterances, which hurts the performance of unsupervised semantic paring. 
In this section, we describe how to alleviate this bias via adaptive fine-tuning. 
Given a text generation model, after pretraining it using paraphrase corpus, we fine-tune it using synthesized $\left<\right.$\texttt{sentence}, \texttt{canonical utterance}$\left.\right>$ pairs.
Previous studies have shown that the pre-training on synthesized data can significantly improve the performance of semantic parsing \citep{XuDq:2020,Marzoev:2020,YuT:2020,Silei:2020}. 
Specifically, 
we design three data synthesis algorithms:
\\
1) \textbf{CUs} 
We sample CUs from SCFGs, and preserve executable ones. 
As we do not have the paired sentences,
we only fine-tune the language model of the PLMs on CUs. \\
2) \textbf{Self Paras} We use the trained paraphrase model to get the natural language paraphrases of the sampled canonical utterances to form $\left<\right.$\texttt{sentence}, \texttt{canonical utterance}$\left.\right>$ pairs. \\
3) \textbf{External Paras} 
We also use external paraphrase methods such as back translation to get the
pairs.




\section{Utterance Reranking}


Adaptive fine-tuning resolves the style bias problem by fitting a better paraphrase model. In this section, we propose an utterance reranking algorithm to further alleviate  the style bias by reranking and selecting the best canonical form.

Given the utterance  $x$ and top-$N$ parsing results $(y_n,c_n), n= 1,2,...,N$, we rerank all candidates by focusing on semantic similarities between $x$ and $c_n$, so that canonical utterances  can be effectively selected. Reranking for semantic parsing has been exploited in many previous studies \citep{Berant:2014,Yin:2019}. These works employ reranking for canonical utterances selection. Differently, our re-ranker does not need labeled data. Formally, we measure two similarities between $x$ and $c_n$ and 
the final reranking score is calculated by:
\begin{equation}
  \label{Equ:score}
  \begin{aligned}
  score(x,c) = & \log p(c|x) +  s_{rec}(x,c)\\ 
           & + s_{asso} (x,c) 
  \end{aligned}
\end{equation}


\paragraph{Reconstruction Score} 
The reconstruction score measures the coherence and adequacy of the canonical utterances, using the probability of reproducing the original input sentence $x$ from $c$ with the trained paraphrasing model:
$s_{rec}(x,c)= \log p_{pr}(x|c)$

\paragraph{Association Score} 
The association score measures whether $x$ and $c$ contain words that are likely to be paraphrases. We calculate it as:

\begin{small}
\begin{equation}
  \begin{aligned}
  s_{asso} (x,c)= & \log{
  \prod_{i=1}^{|c|} \sum_{j=0}^{|x|} p\left(c_{i}|x_{j}\right)a(j|i)
  } \\
  & + \log{
  \prod_{j=1}^{|x|} \sum_{i=0}^{|c|} p\left(x_{j}|c_{i}\right)a(i|j)
  }
  \end{aligned}
\end{equation}
\end{small}
in which, $p\left(c_{i}|x_{j}\right)$ means the paraphrase probability from $x_{j}$ to $c_{i}$, and $a(j|i)$ means the alignment probability. 
The paraphrase probability and alignment are trained and inferred  as the translation model in SMT IBM model 2.

\section{Experiments}


\subsection{Experimental Settings}

\paragraph{Datasets}

We conduct experiments on three datasets: \textsc{Overnight}($\lambda$-DCS), \textsc{Geo}(FunQL), and \textsc{GeoGranno}, which use different meaning representations and on different domains. Our implementations are public available$\footnote{\text{https://github.com/lingowu/ssd}}$.

\indent \textbf{\textsc{Overnight}}\indent  
This is a multi-domain dataset, which contains natural language paraphrases paired with lambda DCS logical forms across eight domains. We  use the same train/test splits as \citet{Wang:2015}.

\indent \textbf{\textsc{Geo}(FunQL)}\indent  
This is a semantic parsing benchmark about U.S.  geography \citep{Zelle:1996} using the variable-free semantic representation FunQL \citep{Kate:2005}. 
We extend the FunQL grammar to SCFG for this dataset.
We follow the standard 600/280 train/test splits.

\indent \textbf{\textsc{GeoGranno}}\indent  
This is another version of \textsc{Geo} \citep{Herzig:2019}, in which lambda DCS logical forms paired with canonical utterances are produced from SCFG.
Instead of paraphrasing sentences, crowd workers are required to select the correct canonical utterance from candidate list.
We follow the split (train/valid/test 487/59/278) in original paper.

\begin{table*}[!t]
  \centering
  \resizebox{0.9\textwidth}{!}{
    \begin{tabular}{l|l|cccccccc|l}
    \Xhline{0.88pt}
    \multicolumn{2}{c|}{   }&
    \bf{Bas.} & \bf{Blo.} & \bf{Cal.} & \bf{Hou.} & \bf{Pub.} & \bf{Rec.} & \bf{Res.} & \bf{Soc.} & \bf{Avg.} \\
    \Xhline{0.88pt}
    \multicolumn{10}{l}{ \textbf{Supervised} } \\
    \hline
    \multicolumn{2}{l|}{\textsc{Recombination} {\cite{Robin:2016}}} & 
85.2 & 58.1 & 78.0 & 71.4 & 76.4 & 79.6 & 76.2 & 81.4 & 75.8 \\
    \multicolumn{2}{l|}{\textsc{CrossDomain} {\citep{Su:2017}}} & 86.2 & 60.2 & 79.8 & 71.4 & 78.9 & 84.7 & 81.6 & 82.9 & 78.2 \\
    \multicolumn{2}{l|}{\textsc{\textsc{Seq2Action}} {\citep{Chen:2018}}} & 
    88.2 & 61.4 & 81.5 & 74.1 & 80.7 & 82.9 & 80.7 & 82.1 & 79.0 \\
    \multicolumn{2}{l|}{\textsc{Dual} {\citep{Cao:2019}}} & 87.5  & 63.7 & 79.8 & 73.0 & 81.4 & 81.5 & 81.6 & 83.0 & 78.9 \\
    \multicolumn{2}{l|}{\textsc{Two-stage} {\citep{Cao:2020}}} & 87.2 & 65.7 & 80.4 & 75.7 & 80.1 & 86.1 & 82.8 & 82.7 & \textbf{80.1} \\
    \hline
    \multicolumn{2}{l|}{SSD (Word-Level)} & 86.2 & 64.9 & 81.7 & 72.7 & 82.3 & 81.7 & 81.5 & 82.7 & 79.2 \\
    \multicolumn{2}{l|}{SSD (Grammar-Level)} & 86.2 & 64.9 & 81.7 & 72.7 & 82.3 & 81.7 & 81.5 & 82.7 & 79.0 \\
    \Xhline{0.88pt}
    \multicolumn{10}{l}{ \textbf{Unsupervised (with nonparallel data)} } \\
    \hline
    \multicolumn{2}{l|}{\textsc{Two-stage} {\citep{Cao:2020}}}
     & 64.7 & 53.4 & 58.3 & 59.3 & 60.3 & 68.1 & 73.2 & 48.4 & 60.7 \\
    \multicolumn{2}{l|}{\textsc{WmdSamples} \citep{Cao:2020}} &
    31.9 & 29.0 & 36.1 & 47.9 & 34.2 & 41.0 & 53.8 & 35.8 & 38.7  \\
    \hline
    \multicolumn{2}{l|}{\textsc{SSD-Samples}  (Word-Level)}
    & 71.7 &  58.7 & 60.1 & 61.7 & 57.6 & 64.3 & 70.9 & 46.0 & 61.4 \\
    \multicolumn{2}{l|}{\textsc{SSD-Samples}  (Grammar-Level)}
     & 71.3 &  58.8 & 60.6 & 62.2 & 58.8 & 65.4 & 71.1 & 49.1 & \textbf{62.2} \\
    \Xhline{0.88pt}
    \multicolumn{10}{l}{ \textbf{Unsupervised} } \\
    \hline
    \multicolumn{2}{l|}{Cross-domain Zero Shot}
    &  - & 28.3 & 53.6 & 52.4 & 55.3 & 60.2 & 61.7 & -  & -    \\
    \multicolumn{2}{l|}{\textsc{GenOvernight}}
    &  15.6 & 27.7 & 17.3 & 45.9 & 46.7 & 26.3 & 61.3 & 9.7  & 31.3    \\
    \multicolumn{2}{l|}{\textsc{Synth-Seq2Seq}}
    & 16.1 & 23.6 & 16.1 & 30.2 & 36.6 & 26.9 & 43.1 & 9.2 & 25.2  \\
    \multicolumn{2}{l|}{\textsc{SynthPara-Seq2Seq}}
    & 28.4 & 37.3 & 33.9 & 38.1 & 39.1 & 41.7 & 62.7 & 23.3 & 38.1 \\
    \hline
    \multicolumn{2}{l|}{\textsc{SSD}  (Word-Level)}
     & 68.3 & 54.9 & 51.2 & 55.0 & 54.7 & 60.2 & 65.4 & 33.6 & 55.4\\ 
    \multicolumn{2}{l|}{\textsc{SSD}  (Grammar-Level)}
     & 68.8 & 58.1 & 56.5 & 56.1 & 57.8 & 59.3 & 66.9 & 37.1 & \textbf{57.6}\\ 
    \Xhline{1pt}
    \end{tabular} }%
    \setlength{\belowcaptionskip}{-0.3cm}
  \caption{Overall results on \textsc{Overnight}.
  }
  \label{Overnight}%
\end{table*}%

\paragraph{Paraphrase Model}
We obtain the paraphrase model by training T5 and GPT2.0 on WikiAnswer Paraphrase$\footnote{\text{http://knowitall.cs.washington.edu/ paralex}}$, we train 10 epochs 
with learning rate as 1e-5. Follow \citet{DecomNPG:2019}, we sample 500K pairs of sentences in WikiAnswer corpus as training set and 6K as dev set. 
We generate adaptive fine-tuning datasets proportional to their labeled datasets, 
and back-translation(from English to Chinese then translate back) is used to obtain external paraphrases data. 
On average, we sample 423 CUs per domain, and synthesize 847 instances per domain in Self Paras and 1252 in External Paras.




\paragraph{Unsupervised settings} 
In unsupervised settings, we do not use any annotated semantic parsing data. 
The paraphrase generation models are fixed after the paraphrasing pre-training and the adaptive fine-tuning. The models are employed to generate canonical utterances and MRs synchronously via rule-level or word-level inference. 
In rule-level inference, the leftmost non-terminators are eliminated by cyclically expanded and the maximum depth $K$ is set to 5, the beam size is set to 20. 
\textsc{SSD} uses T5 as the pre-trained language model in all the proposed components, including adaptive fine-tuning, reranking and the two decoding constraints. Ablation experiments are conducted over all components with rule-level inference.

\begin{table*}[h]
  \centering
  \resizebox{0.97\textwidth}{!}{
    \begin{tabular}{lccccccccccc}
    \toprule
    & \bf{Bas.} & \bf{Blo.} & \bf{Cal.} & \bf{Hou.} & \bf{Pub.} & \bf{Rec.} & \bf{Res.} & \bf{Soc.} 
    & 
    \renewcommand\arraystretch{0.8}
    \tabincell{l}{\fontsize{10pt}{10pt}{\bf{\textsc{Overn.}}}\\\fontsize{10pt}{10pt}{\bf{Avg.}}}  
    &
    \renewcommand\arraystretch{0.8}
    \tabincell{l}{\fontsize{10pt}{10pt}{\bf{\textsc{Geo}}}\\\fontsize{10pt}{10pt}{\bf{\textsc{Granno}}}} 
    & 
    \renewcommand\arraystretch{0.8}
    \tabincell{l}{\fontsize{10pt}{10pt}{\bf{\textsc{Geo}}}\\\fontsize{10pt}{10pt}{\bf{(FunQL)}}} \\
    \hline    
    \specialrule{0em}{0pt}{1pt}
    \textsc{CompleteModel}   
    & 68.8 & 58.1 & 56.5 & 56.1 & 57.8 & 59.3 & 66.9 & 37.1 & \textbf{57.6} & \bf{58.5} & \bf{63.2}\\ 
    \hline
    \specialrule{0em}{0pt}{1pt}
    \multicolumn{12}{l}{ \textbf{Constraints} } \\
    \hline
    $\ \ $  - \textsc{Semantic} & 65.5 & 54.4 & 52.4 & 51.9 & 56.5 & 57.4 & 62.0 & 35.7 & 54.5 & 56.5 & 61.1 \\
    $\ \ $ - \textsc{Grammar}  & 63.9 & 52.1 & 47.6 & 50.3 & 45.3 & 52.3 & 56.3 & 29.9 &  
    49.7 &  50.7 & 53.9\\ 
    \hline
    \specialrule{0em}{0pt}{1pt}
    \multicolumn{12}{l}{ \textbf{Adaptive Fine-tuning} } \\
    \hline
    $\ \ $ - \textsc{External Paras} & 66.8 & 56.6 & 51.6 & 54.0 & 48.4 & 56.9 & 63.3 & 32.4 & 
    53.8 &   56.5 & 61.1  \\ 
    $\ \ $ - \textsc{Paras} (Only CUs) & 64.5 & 55.4 & 50.0 & 51.3 & 47.8 & 55.6 & 62.0 & 31.7 & 
    52.3 &  54.3 & 59.6 \\ 
    $\ \ $ - \textsc{Fine-tuning} & 63.9 & 53.6 & 48.4 & 47.6 & 44.9 & 53.2 & 62.5 & 31.4 & 
    50.7 & 51.8 & 55.4   \\ 
    \hline
    \specialrule{0em}{0pt}{1pt}
    \multicolumn{12}{l}{ \textbf{Reranking} } \\
    \hline
    $\ \ $ - \textsc{Reranking} & 58.2 & 57.2 & 56.3 & 50.2 & 55.7 & 58.0 & 62.9 & 37.7  & 
    54.6 &  57.2 &  62.5 \\ 
    $\ \ $ \textsc{Oracle} (R@20) & 71.2 & 83.2 & 86.9 & 76.9 & 82.7 & 86.9 & 88.1 & 62.8 & 
    79.8 &  70.9 &  81.4 \\ 
    \hline
    \specialrule{0em}{0pt}{1pt}
    \multicolumn{12}{l}{ \textbf{Pretrained Language Models} } \\
    \hline
    $\ \ $ GPT-2 & 67.0 & 54.6 & 53.7 & 55.7 & 56.1 & 58.9 & 66.4 & 32.7 & 
    55.6 &  58.3 &  62.1 \\ 
    $\ \ $ \textsc{Rand} & 58.3 & 48.6 & 45.8 & 47.6 & 50.3 & 55.6 & 54.5 & 30.3 & 
    48.9 &   51.4 &  54.3 \\ 
    \bottomrule
    \end{tabular}  }%
    \setlength{\abovecaptionskip}{0.1cm}
    \setlength{\belowcaptionskip}{-0.2cm}
  \caption{ Albation results of our model with different settings on the three datasets.
  }
  \label{abaltion_all}%
\end{table*}%

\paragraph{Unsupervised settings (with external nonparallel data)}
\citet{Cao:2020} have shown that external nonparallel data (including nonparallel  natural language utterances and canonical utterances) can be used to build unsupervised semantic parsers. For fair comparison, we also conduct unsupervised experiments with external unparallel data. Specifically, we enhance the original SSD using the \textsc{Samples} methods \citep{Cao:2020}: we label each input sentences with the most possible outputs in the nonparallel corpus and use these samples as peusdo training data – we denote this setting as \textsc{SSD-Samples}.

\paragraph{Supervised settings}
Our SSD method can be further enhanced using annotated training instances. Specifically, given the annotated $\left<\text{utterance, logical form}\right>$ instances, we first transform logical form to its canonical form, then use them to  further fine-tune  our paraphrase models after unsupervised pre-training. 

\paragraph{Baselines} 
We compare our method with the following unsupervised baselines: 1) Cross-domain Zero Shot\citep{Herzig:2018}, which trains on other source domains and then generalizes to target domains in \textsc{Overnight} and 2) \textsc{GenOvernight}\citep{Wang:2015} in which models are trained on synthesized $\left<\text{CU, MR}\right>$ pairs; 3) We also implement \textsc{Seq2Seq} baseline on the synthesized data as \textsc{Synth-Seq2Seq}. 4) \textsc{SynthPara-Seq2Seq} is trained on the synthesized data and $\left<\text{CU paraphrase, MR}\right>$ pairs, the paraphrases are obtained in the same way in Section \ref{adaptive}.

\subsection{Experimental Results}

\subsubsection{Overall Results}

\begin{table}[!t]
  \centering
  \resizebox{0.48\textwidth}{!}{
    \begin{tabular}{|l|l|c|c|}
    \hline
    \multicolumn{2}{|c|}{   }
    &
    \renewcommand\arraystretch{0.8}
    \tabincell{l}{\specialrule{0em}{1pt}{1pt}\fontsize{10pt}{10pt}{\bf{\textsc{Geo}}}\\\fontsize{10pt}{10pt}{\bf{\textsc{Granno}}}} 
    & 
    \renewcommand\arraystretch{0.8}
    \tabincell{l}{\specialrule{0em}{1pt}{1pt}\fontsize{10pt}{10pt}{\bf{\textsc{Geo}}}\\\fontsize{10pt}{10pt}{\bf{(FunQL)}}} \\
    \hline
    \multicolumn{4}{|c|}{ \textbf{Supervised} } \\
    \hline
    \multicolumn{2}{|l|}{\textsc{DepHT} \small{\citep{Zhanming:2018}}} & - & 89.3 \\
    \multicolumn{2}{|l|}{\textsc{CopyNet} \small{\citep{Herzig:2019}}} & 72.0 & - \\
    \multicolumn{2}{|l|}{One-stage \small{\citep{Cao:2020}}} & 71.9 & - \\
    \multicolumn{2}{|l|}{Two-stage \small{\citep{Cao:2020}}} & 71.6 & - \\
    \multicolumn{2}{|l|}{\textsc{Seq2Seq} \small{\citep{Guo:2020}}} & - & 87.1 \\
    \hline
    \multicolumn{2}{|l|}{\textsc{SSD }(Word-Level)} &  72.9  &  88.3 \\
    \multicolumn{2}{|l|}{\textsc{SSD }(Grammar-Level)} & 72.0 & 87.9 \\
    \hline
    \multicolumn{4}{|c|}{ \textbf{Unsupervised (with nonparallel data)} } \\
    \hline
    \multicolumn{2}{|l|}{Two-stage \small{\citep{Cao:2020}}} & 63.7  & - \\
    \multicolumn{2}{|l|}{\textsc{WmdSamples} \small\citep{Cao:2020}} &  35.3 & -\\
    \hline
    \multicolumn{2}{|l|}{\textsc{SSD-Samples} (Word-Level)} &  64.0 & 64.3 \\
    \multicolumn{2}{|l|}{\textsc{SSD-Samples} (Grammar-Level)} &  \textbf{64.4} & \textbf{65.0}\\
    \hline
    \multicolumn{4}{|c|}{ \textbf{Unsupervised} } \\
    \hline
    \multicolumn{2}{|l|}{\textsc{Synth-Seq2Seq}} &  32.7  &   36.1 \\
    \multicolumn{2}{|l|}{\textsc{SynthPara-Seq2Seq}} &  41.4 &  45.4 \\
    \hline
    \multicolumn{2}{|l|}{\textsc{SSD} (Word-Level)} &  57.2 & 62.8  \\
    \multicolumn{2}{|l|}{\textsc{SSD} (Grammar-Level)} &  \textbf{58.5 }& \textbf{63.2 }\\
    \hline
    \end{tabular} }%
    \setlength{\abovecaptionskip}{0.1cm}
    \setlength{\belowcaptionskip}{-0.3cm}
  \caption{Overall results on \textsc{GeoGranno} and \textsc{Geo}(FunQL).
  }
  \label{Geo}%
\end{table}%


The overall results of different baselines and our method are shown in Table \ref{Overnight} and Table \ref{Geo} (We also demonstrate several cases in Appendix). For our method, we report its performances on three settings. We can see that:


1. \textbf{By synchronously decoding canonical utterances and meaning representations, SSD achieves competitive unsupervised semantic parsing performance.} In all datasets, our method outperforms other baselines in the unsupervised settings. These results demonstrate that unsupervised semantic parsers can be effectively built by simultaneously exploit semantic and structural constraints, without the need of labeled data.


2. \textbf{Our model can achieve competitive performance on different datasets with different settings.}  In supervised settings, our model can achieve competitive performance with SOTA. With nonparallel data, our model can outperform Two-stage. 
On \textsc{Geo}(FunQL) our model also obtains a significant improvement compared with baselines, 
which also verifies that our method is not limited to specific datasets (i.e., \textsc{Overnight} and \textsc{GeoGranno}, which are constructed with SCFG and paraphrasing.)

3. \textbf{Both rule-level inference and word-level inference can effectively generate paraphrases under the grammar constraints.} The rule-level inference can achieve better performance, we believe this is because rule-level inference is more compact than word-level inference, 
therefore the rule-level inference can search wider space and benefit beam search more.

\subsubsection{Detailed Analysis}



\paragraph{Effect of Decoding Constraints}
To analyze the effect of decoding constraints, we conduct ablation experiments with different constraint settings and the results are shown in Table \ref{abaltion_all}: \textsc{-Semantic} denotes removing the semantic constraint, \textsc{-Grammar} denotes all constraints are removed at the same time, the decoding is unrestricted. We can see that the constrained decoding is critical for our paraphrasing-based semantic parsing, and both grammar constraints and semantic constraints contribute to the improvement.

\paragraph{Effect of Adaptive Fine-tuning}
To analyze the effect of adaptive fine-tuning, we show the results with different settings by ablating a fine-tuning corpus at a time (see Table \ref{abaltion_all}). We can see that adaptive fine-tuning can significantly improve the performance. And the paraphrase generation model can be effectively fine-tuned only using CUs or Self Paras, which can be easily constructed.


\paragraph{Effect of Reranking}
To analyze the effect of reranking, we compare the settings with/without reranking and its upper bound -- Oracle, which can always select the correct logical form if it is within the beam. 
Experimental results show that reranking can improve the semantic parsing performance. 
Moreover, there is still a large margin between our method and Oracle, i.e., the unsupervised semantic parsing can be significantly promoted by designing better reranking algorithms.

\begin{figure}[t]
  \centering
  \noindent\makebox[.419\textwidth][r] {
  \includegraphics[width=.42\textwidth]{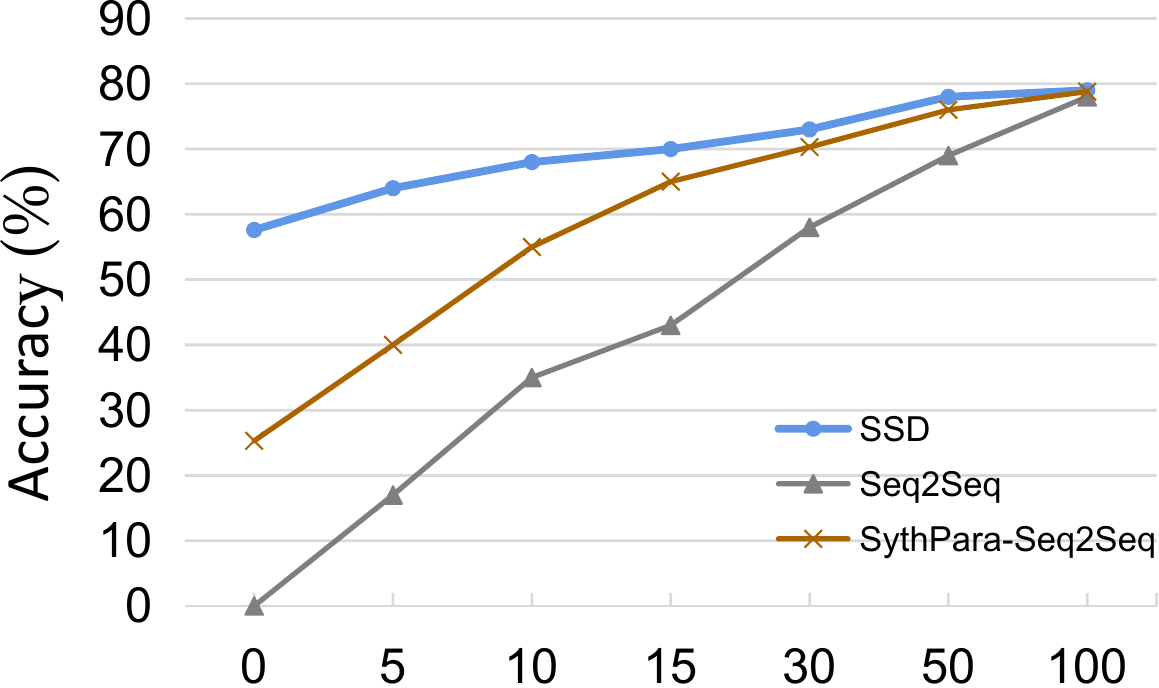} 
  }
    \setlength{\abovecaptionskip}{0.1cm}
    \setlength{\belowcaptionskip}{-0.1cm}
  \caption{Semi-supervised results of different ratios of labeled data on \textsc{Overnight}. }
  \label{semi} 
\end{figure}

\paragraph{Effect of Adding Labeled Data}
To investigate the effect of adding labeled data, we test our method by varying the size of the labeled data on \textsc{Overnight} from 0\% to 100\%. In Fig. \ref{semi}, we can see that our method can outperform baselines using the same labeled data. And a small amount of data can produce a good performance using our method.

\paragraph{Effect of Pretrained Language Models}
To analyze the effect of PLMs, we show the results with different PLM settings: 
instead of T5 we use GPT-2 or randomly initialized transformers to construct paraphrasing models.
Experimental results show that powerful PLMs can improve the performance. 
Powered by the language generation models to do semantic parsing,
 our method can benefit from the rapid development of PLMs.

\section{Related Work}



\paragraph{Data Scarcity in Semantic Parsing.}
Witnessed the labeled data bottleneck problem, many techniques have been proposed to reduce the demand for labeled logical forms.
Many weakly supervised learning are proposed \citep{Artzi:2013,Berant:2013,Reddy:2014,Agrawal:2019,Macro-action/Chen:2020}, 
such as denotation-base learning \citep{Liang:2016,Goldman:2018}, iterative searching \citep{Dasigi:2019}.
Semi-supervised semantic parsing is also proposed\citep{Lexicon/Chen:2018}. 
Such as variational auto-encoding
\citep{Yin:2018},
dual learning framework for 
semantic parsing
\citep{Cao:2019},
dual information maximization method 
\citep{Ye:2019},
and back-translation \citep{SunTang:2019}.
One other strategy is to generate data for semantic parsing, e.g.,
\citet{Wang:2015} construct a semantic parsing dataset from grammar rules and crowdsourcing paraphrase. 
\citet{Guo:2018} produce pseudo-labeled data.
\citet{Robin:2016} create new ``recombinant'' training examples with SCFG.
The domain transfer techniques are also used to reduce the cost of data collecting for the unseen domain \citep{Su:2017,Herzig:2018,LuArab:2019,Zhong:2020}.
\citet{Goldwasser:2011,Poon:2009,Schmitt:2020} leverage external resources or techniques for unsupervised learning.
\paragraph{Constrained Decoding.} After neural parsers model semantic parsing as a sentence to logical form translation task \citep{Yih:2015,Krishnamurthy:2017,Iyyer:2017,Zhanming:2018,Lindemann:2020}, 
many constrained decoding algorithms are also proposed, such as 
type constraint-based illegal token filtering \citep{Krishnamurthy:2017}; 
Lisp interpreter-based method \citep{Liang:2017}; 
type constraints for generating valid actions \citep{Iyyer:2017}.

\paragraph{Paraphrasing in Semantic Parsing.} 
Paraphrase models have been widely used in semantic parsing.
{ParaSempre} \citep{Berant:2014} use paraphrase model to rerank candidate logical forms.
\citet{Wang:2015} employ SCFG grammar rules to produce MR and canonical utterance pairs, and construct \textsc{Overnight} dataset by paraphrasing utterances. 
\citet{Dong:2017} use paraphrasing to expand the expressions of query sentences.
Compared with these methods, we combine paraphrasing with grammar-constrained decoding, therefore SSD can further reduce the requirement of labeled data and achieve unsupervised semantic parsing.


\section{Conclusions}

We propose an unsupervised semantic parsing method -- Synchronous Semantic Decoding, 
which leverages paraphrasing and grammar-constrained decoding  
to simultaneously resolve the semantic gap and the structure gap. 
Specifically, we design two synchronous semantic decoding algorithms for paraphrasing under grammar constraints, 
and exploit adaptive fine-tuning and utterance reranking to alleviate the style bias in semantic parsing. 
Experimental results show that our approach can achieve competitive performance in unsupervised settings.

\section*{Acknowledgments}
We sincerely thank the reviewers for their insightful comments and valuable suggestions. Moreover, this work is supported by the National Key Research and Development Program of China(No. 2020AAA0106400),  the National Natural Science Foundation of China under Grants no. 61906182 and 62076233, and in part by the Youth Innovation Promotion Association CAS(2018141).

\bibliography{SSD}
\bibliographystyle{acl_natbib}
\appendix
  \renewcommand{\appendixname}{Appendix~\Alph{section}}

\begin{table}[!b]
  \centering
  \hspace*{0.05\textwidth}
  \resizebox{0.9\textwidth}{!}{
    \begin{tabular}{p{12cm}|c|c|c|c}
    \toprule
    \textbf{Cases} & 
    \tabincell{c}{$x \rightarrow c$\\$log P(c|x)$}
     & 
    \tabincell{c}{$x \leftarrow c$\\$s_{rec}(x,c)$ }
     & 
    \tabincell{c}{$x \leftrightarrow c$\\$s_{asso}(x,c)$ }
     &  
     \renewcommand\arraystretch{0.6}
    \tabincell{c}{\small{Overall} \\\small{Reranking} \\ \small{Score}}
       \\
    \midrule
    \textbf{Input:} restaurants that accept credit cards and offer takeout &&&  \\
    \textbf{Outputs:} \underline{restaurant that takes credit cards and that has take-out} & -54.2  & -3.1& -20.3 & \textbf{-77.6} \\
    \textcolor{white}{\textbf{Outputs:}} \underline{restaurant that has take-out and that takes credit cards}& -72.0 & -8.8  & -20.3& -101.1\\
    \textcolor{white}{\textbf{Outputs:}} restaurant that takes credit cards & -77.2  & -22.4 & -31.9 & -131.5\\
    \textcolor{white}{\textbf{Outputs:}} restaurant that takes credit cards and that takes credit cards & -84.2  & -26.7 & -28.1 & -139.0 \\
    \midrule
    \textbf{Input:} meetings held in the same place as the weekly standup meeting &&&  \\
    \textbf{Outputs:} meeting whose date is date of weekly standup & -62.2  & -40.2 & -67.1 & -169.5\\
    \textcolor{white}{\textbf{Outputs:}}  \underline{meeting whose location is location of weekly standup} & -62.7  & -22.1 &  -62.4& \textbf{-147.2}\\
    \textcolor{white}{\textbf{Outputs:}} \underline{meeting whose location is location that is location of weekly standup}& -65.0 & -21.1  & -63.5 & 149.6\\
    \textcolor{white}{\textbf{Outputs:}} meeting whose date is at most date of weekly standup & -67.2  & -35.8 & -73.3 & -176.3\\
    \midrule
    \textbf{Input:} meetings held in the same place as the weekly standup meeting &&&  \\
    \textbf{Outputs:}  \underline{meeting whose location is location of weekly standup} & -2.7  & -22.1 & -62.4 & \textbf{-86.2}\\
    \textcolor{white}{\textbf{Outputs:}} \underline{meeting whose location is location that is location of weekly standup}& -6.0 & -21.1  & -63.5 & -90.6 \\
    \textcolor{white}{\textbf{Outputs:}} location that is location of weekly standup & -18.0 & -32.6  & -62.8 & -113.4 \\
    \textcolor{white}{\textbf{Outputs:}} meeting whose date is date of weekly standup & -31.2  & -40.2 & -67.1 & -138.5 \\
    \bottomrule
    \end{tabular}   
    }%

  \label{without-fine-tuning}%
\end{table}%

\begin{table}[!b]
  \centering
  \vspace*{-2pt}
  \resizebox{1.013\textwidth}{!}{

  \hspace*{-0.0225\textwidth}
    \begin{tabular}{p{15.89cm}}
    Table 4: Output cases from SSD on \textsc{Overnight}.
    The outputs are sorted by the generation score $log P(c|x)$. The underlined canonical utterances are correct. Adaptive fine-tuning is employed in the third case, and not employed in the first two cases.
  Accuracies on \textsc{Overnight}. 
  In \textsc{GenOvernight} \citet{Wang:2015}, all the canonical utterances are also generated without manual annotation. 
  The previous methods with superscript $*$ means they use different unsupervised settings.



       \\
    \end{tabular}   

    }%

  \label{without-fine-tuning}%
\end{table}%

\section{Appendix}

\subsection{Case Study}
In Table 4, we present the cases generated from SSD. 
Cases show that SSD can output semantics-similar and grammar-legal utterances. In case 1, ``take-out'' does not appear in paraphrase dataset, we can still efficiently generate the utterances containing it, which shows our constrained-paraphrasing based semantic parser has the generalization ability on unseen words.
We found that the parser maintains high recall, covering the correct canonical utterances in our n-best list of predictions.
As case 2 shows the designed utterance reranking score can select the best canonical utterances by focusing on coherence and adequacy.
With adaptive fine-tuning (case 3), our model can generate the utterances focusing more on semantics to alleviate the style bias.

\end{document}